\documentclass[]{spie}  %>>> use for US letter paper
\usepackage{amsmath,amsfonts,amssymb}
\usepackage{graphicx}
\usepackage[colorlinks=true, allcolors=blue]{hyperref}

\title{Multi-Object Tracking with Deep Learning Ensemble for Unmanned Aerial System Applications}

\author[a]{Wanlin Xie}
\author[a]{Jaime Ide}
\author[a]{Daniel Izadi}
\author[a]{Sean Banger}
\author[a]{Thayne Walker}
\author[a]{Ryan Ceresani}
\author[a]{Dylan Spagnuolo}
\author[a]{Christopher Guagliano}
\author[a]{Henry Diaz}
\author[a]{Jason Twedt}

\affil[a]{Lockheed Martin AI Center}

\authorinfo{Further author information: (Send correspondence to J.T.)\\
J.T.: E-mail: jason.c.twedt@lmco.com}

\begin{document}
\maketitle

\begin{abstract}
Multi-object tracking (MOT) is a crucial component of situational awareness in
military defense applications. With the growing use of unmanned aerial systems
(UASs), MOT methods for aerial surveillance is in high demand. Application of
MOT in UAS presents specific challenges such as moving sensor, changing zoom
levels, dynamic background, illumination changes, obscurations and small
objects. In this work, we present a robust object tracking architecture aimed to
accommodate for the noise in real-time situations. We propose a
kinematic prediction model, called Deep Extended Kalman Filter (DeepEKF), in which
a sequence-to-sequence architecture is used to predict entity trajectories in
latent space. DeepEKF utilizes a learned image embedding along with an attention
mechanism trained to weight the importance of areas in an image to predict future states.
For the visual scoring, we experiment with different similarity measures to
calculate distance based on entity appearances, including a convolutional neural
network (CNN) encoder, pre-trained using Siamese networks. In initial evaluation
experiments, we show that our method, combining scoring structure of the
kinematic and visual models within a MHT framework, has improved performance
especially in edge cases where entity motion is unpredictable, or the data
presents frames with significant gaps.
\end{abstract}

\keywords{multi-object tracking, deep neural network, extended kalman filter,
long short-term memory network, multiple hypothesis tracking}

\section{INTRODUCTION}
Unmanned aerial systems (UAS) are commonly used for intelligence surveillance
and reconnaissance (ISR) missions to provide critical information to decision
makers. Situational awareness (SA) is enhanced when up-to-date, coherent, and trustworthy information is disseminated to all areas
of the battlefield. Artificial Intelligence and Machine Learning (AI/ML)
deployed at the edge is a key enabler to achieving this. Edge compute devices
provide the infrastructure for deep learning methods to accelerate information
extraction from onboard sensors and disseminate over communication networks.
A full understanding of the battlefield requires all entities and their intent
be known and understood. Multi-object tracking (MOT) is a crucial component to
achieving this \cite{Asnis2011, Blackman2013, Shi2020}. Application of MOT in
UAS presents specific challenges such as moving platform and sensor,
large variations in view point and pose, fast and abrupt object motion, changing
zoom levels, illumination changes, occlusions, low resolution of detected objects
due to small sizes \cite{Li_Yeung_2017,Du2018}.

\begin{figure}[h]
   \begin{center}
         \includegraphics[scale=0.75]{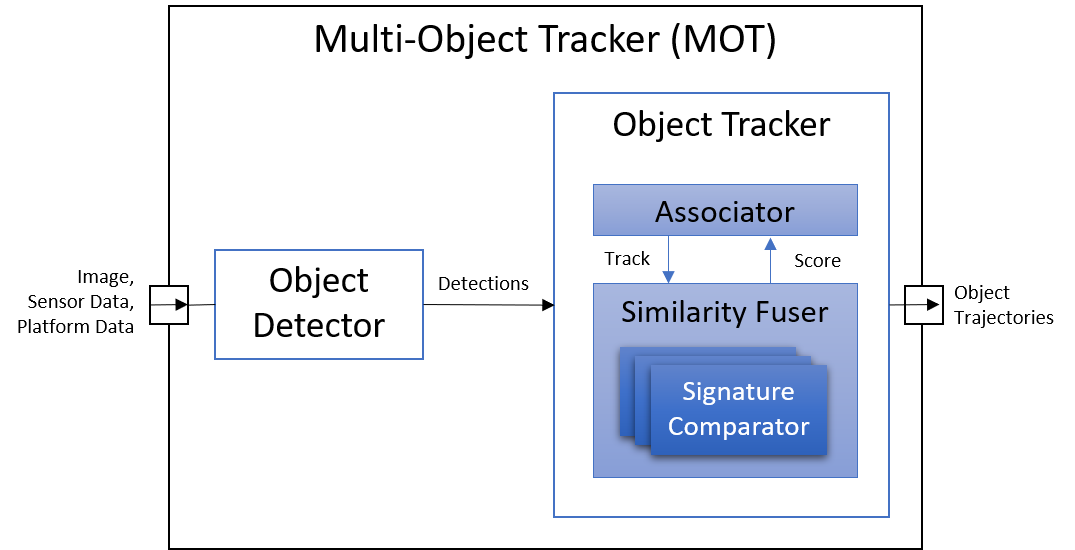}
         \caption{MOT Functional Architecture}
         \label{fig:architecture}
   \end{center}
\end{figure}

\newpage
Our solution is based on the tracking-by-detection paradigm
\cite{Andriluka2008,Porikli2012,Feichtenhofer2017}. As shown in Figure \ref{fig:architecture}, platform information, images and other sensor data are received by the multi-object tracker (MOT). Objects are identified as \emph{Detections} by the Object Detector. A Detection consists of a bounding box on each object entity as well as the corresponding entity label. Detections are then sent to the \emph{Object Tracker}, which is composed of an \emph{Associator} and \emph{Similarity Fuser}. The job of the Associator is to link together detections between frames to create object trajectories. The Similarity fuser is a modular scoring component that compiles all the scores from the \emph{Signature Comparators} into a final metric.

Track performance is highly dependent on the similarity computation between detections
and/or \emph{tracklets} with both motion and appearance methods playing a significant
role \cite{Ciaparrone2020}. To this end, we propose an architecture that is composed
of an ensemble of Signature Comparators, each targeting different aspects of tracking.
We broadly separate these Signature Comparators into two categories, short-term and long-term.
In both cases, we aim to exploit the uniqueness of an object through detection
of intrinsic signatures (fingerprint) using deep learning. \emph{short-term}
methods focus on modeling the motion of objects in subsequent frames over short
durations. Here an object's anticipated motion or kinematic signature is
exploited to associate the object in future frames. Over periods of long
occlusions or as objects move in and out of the sensor field-of-view (FOV),
motion based models can become less accurate due to error propagation over time.
In these cases we exploit an object's visual signature for re-identification of
the object in future frames (\emph{long-term} tracking). The output of all
Signature Comparators are combined into a single score by the Similarity Fuser and
provided to the Associator for track ID association. In this work, we combine
recent advances in deep learning \cite{Zhao2019,Xu2020} with traditional object
tracking methods \cite{Pylyshyn1988TrackingMI,Chong2018:MHT40}.

We implement the Associator as a tree-based multiple hypothesis tracking (MHT)
\cite{Reid1979, KimRehg2015:MHT} framework that produces the best hypothesis
based on the kinematic and visual information extracted from the detections.
MHT provides a robust tracking system to deal with misdetections and occlusions
by keeping multiple trees of plausible object trajectories. However, it heavily
relies on the accuracy of the kinematic and visual scores. Therefore, we propose
improvements in two folds through the addition of both short and long-term
Signature Comparators:

\begin{itemize}
\item{Deep learning the kinematic scores. In MHT, high kinematic scores are
given to objects that are observed close to their predictions. Thus, the
accuracy of the kinematic model will drive the reliability of the kinematic
scores. Traditionally, this motion prediction can be done using a standard Kalman Filter \cite{Kalman1960}. This works well on a simulated scene or when a camera is fixed at a
stationary position, but is unreliable as more real world factors are
introduced, e.g. non-linear motion of the aircraft, sensor pointing angles,
sensor zoom, and object trajectories. In this work, we present a novel model
using deep learning models integrated within an extended Kalman Filter (EKF) \cite{Yang2017}
based Signature Comparator to predict nonlinear and complex movement patterns.
We propose a kinematic prediction model, called DeepEKF, in which we use a
sequence-to-sequence architecture to predict entity trajectories in latent space.
DeepEKF utilizes a learned image embedding along with an attention
mechanism that is trained to weight the importance of areas in an image to
predict future states. The model is composed of a dual deep neural network (DNN) \cite{LeCun2015} - a kinematic prediction model with an Encoder-Decoder Long Short-Term Memory
(LSTM) architecture trained end-to-end with a convolutional neural network (CNN) \cite{Albawi2017}}

\item{Deep learning the visual scores. Along with the kinematic score, the
visual similarity between two detected objects is used to estimate the final
score of the tracking hypothesis in MHT. We employ the sum square difference
(SSD) \cite{KimRehg2015:MHT} as the baseline measure, and propose a deep learning model to estimate
the similarity score between two image chips. The model uses a CNN architecture
and it is trained to learn image embeddings that are optimal in identifying
unique entities. The training is done using Siamese networks \cite{Chopra2005,Taigman2014}.
Our Siamese network architecture is an multi-headed attention-driven model that
is able to focus on specific spatial regions of interest to create a unique object
fingerprint. This fingerprint may be used to perform visual object
re-identification in the absence of reliable kinematic data.}
\end{itemize}

In initial evaluation experiments, we show that our method, combining scoring
structure of the kinematic and visual models within a MHT framework, has
improved performance especially in edge cases where entity motion is
unpredictable, or the data presents frames with significant gaps.

\section{RELATED WORK}
Motion based affinity models predict a future object state and then compare
that state to received detections for similarity scoring. Kalman filters
then predict future bounding box coordinates and generate an affinity score
from intersection over union (IOU) \cite{Wang2017}, and assumes linear motion.
A purely visual approach proposed in \cite{Zhou2019} uses a visual displacement
CNN that takes stacked image chips as input to predict an object's future
location in a discretized frame. Recurrent neural networks (RNNs) have been
proposed to model non-linear motion and measurement noise \cite{Sadeghian2017},
and to overcome the long-term occulsion problem \cite{Babaee2018}. Approaches that use a pure visual approach or the IOU score of detections work well on a stationary camera where the next frame points to the same position, but becomes unreliable when utilized in a scenario with a moving camera.

Visual object re-identification is a rising area of research and much of the
recent work has been improving re-identification using multiple image
comparisons \cite{wieczorek2021unreasonable}, utilizing attention
\cite{liao2021transformerbased}, unsupervised learning \cite{fu2021unsupervised},
body part information fusion \cite{8099877}, and creating clothing agnostic models using generative networks
\cite{ge2018fdgan, li2020learning}. Many open source models are not designed in
the context of long term re-identification (appearance agnostic) and fall short
when challenged with appearance changes. Research that is geared towards appearance agnostic re-identification often
times relies on either generative models or utilizing part detection.
One approach to using generative models is to project an image into the context of
another (pose, color, etc.) \cite{ge2018fdgan} before performing image
comparisons.  This method has the potential to be memory and timing intensive.
Other generative models utilize image reconstruction in a multi-task learning
context to enforce that the feature extractor learns to encode body structure
\cite{li2020learning} before globally pooling the features. Outside of generative models,
utilizing part detection \cite{8099877} guides the visual comparison network to focus on specific
regions of interest in the image that may be generally appearance agnostic.  The part detection
method removes the ability for the model to identify more significant regions
of interest and is not guaranteed to be robust against appearance changes
within the detected parts.  In contrast, our visual
re-identification module uses a spatial attention head to identify
important and unique areas of interest in an image with a method that avoids
global pooling.  Our method supports the use of single and multiple image
comparisons and is adaptable to different viewing angles. Our approach also
does not preclude the inclusion of additional multi-task loss functions that
may prove effective for remaining appearance agnostic.

Work in integrating deep learning techniques in the multiple hypothesis tracking (MHT) framework is growing. Kim et al. \cite{Kim2018} used LSTM networks to model object appearance over time, utilizing a vanilla LSTM to process motion data and bilinear LSTM to process appearance data. CNNs \cite{Chen2017} have also been used in MHT where the CNN was used for feature extraction and combined with a Kalman Filter for motion prediction. Our approach offers the flexibility to combine multiple affinity models in MHT while most approaches have only looked at using a single network. We also introduce a novel deep learning network focused on motion prediction to inform a kinematic likelihood during the MHT gating step.

\section{BACKGROUND}

\subsection{State Estimation}
The Kalman filter (KF) is a popular algorithm for target state estimation from
noisy observations. It has various applications including autonomous vehicles or
stock market prediction. The KF models transition to the next state as a linear
function of the current state. Kalman filters are composed of a two-step
iterative process including a prediction step and an update step that estimates
the mean and variance of a Gaussian distribution. The Extended Kalman Filter
(EKF) models nonlinear state transitions and/or measurement models that are
nonlinear functions of state. The likelihood that a measurement is generated
from a Kalman filter state is given by the multivariate probability density
function \cite{Ramachandra2000}.

An MOT application deployed for aerial ISR missions, must be robust to diverse
operational environments and target sets. The aircraft, sensor pointing angles,
sensor zoom, and targets of interest exhibit complex (nonlinear) behavior that
cannot be specified \emph{a-priori}. A solution that dynamically adapts to
changing state parameters in aircraft, sensor and targets is desired.
Deep learning has shown to be effective in this respect. Learned extensions to
Kalman filters with LSTMs have been shown to improve estimation accuracy
\cite{coskun2017long,Zheng2019}. When applied to pixel-based tracking, LSTMs
show effectiveness in learning the motion of generic objects based on appearance.
\cite{gordon2018re3}. Our approach builds upon prior LSTM motion model
architectures that leverage appearance features and augments them to include
system parameters such as aircraft and sensor state. The network architecture is
modeled after an EKF and includes state prediction, noise estimation and likelihood
calculation within an N-dimensional latent space.

\subsection{General Visual Comparison Methods}
Current visual object tracking models use the method of feature learning,
which creates a feature vector representation of the object being tracked.
These models are trained on ranking loss functions with the objective of
minimizing the agreement between features of dissimilar labels and maximizing
agreement between features of the same label in latent space. Typical metrics for inference of visual feature embeddings are Cosine Similarity
and Euclidean Distances between two embeddings to create a similarity score for
a comparison between two images. These scores can either be ranked against a
gallery of embeddings to find the most likely match, or paired with a margin to
establish whether or not a pair of images is a match.

Many existing models use popular CNN models such as ResNet \cite{He2015}, EfficientNet \cite{Tan2020},
GoogleNet \cite{Szegedy2014}, etc. for the backbone of their re-identification models.
For the purpose of visual feature learning for tracking, there are three typical
approaches to training these models. The first is to use a global feature from
the entire image, and train on classification loss. The second is to use a
ranking loss function to maximize the score between global features of similar
objects. The last is to learn part-based features rather than global features.
Most state-of-the-art (SOTA) models use a combination of a ranking loss function
on both global and local features, and classification loss on the unique ids in
the training set improve generalizability and performance of their models
\cite{chen2020simple, fu2021unsupervised}.

\subsection{Multiple Hypothesis Tracking}
Data association is at the core of multi-object tracking (MOT), and since its
creation, multiple hypothesis tracking (MHT) \cite{Reid1979} became the standard
method for MOT when data association is challenging due to occlusion, low
probability of detection or high target density
\cite{Mori1986,Roy1997:SPIE, Blackman2004,Danchick2006,KimRehg2015:MHT,Brekke2018, Chong2018:MHT40}.
A computationally more efficient alternative to the original hypothesis-oriented
MHT has been proposed and it is known as the tracking-oriented MHT
\cite{Blackman2001:trackOriented, Blackman2004}. As opposed to the original MHT
in which multiple hypotheses are expanded and kept across frames, in
track-oriented MHT, association hypotheses are maintained at the individual
track level. Thus, hypotheses with low probability or that are contradictory are
deleted, and most likely hypothesis tracks are kept.

With the advent of efficient and accurate object detection algorithms provided
by deep learning methods, MHT approaches using the tracking-by-detection framework are popular \cite{KimRehg2015:MHT}. In this approach, objects along with
their bounding boxes are automatically generated, and used as the observations
within MHT. A \emph{track hypothesis} is defined as a sequence of observations
across frames with the same unique identifier. The \emph{global hypothesis} is
defined as the set of track hypotheses that are not in conflict. The key idea in
MHT is to create multiple track trees representing unique objects, and keep them
active across frames, until association ambiguities are resolved. At each frame,
track trees are updated from the observations, scored, and the best set of
non-conflicting tracks (i.e., best global hypothesis) can be found solving a
maximum weighted independent set (MWIS) problem \cite{Papageorgiou2009:MWIS}.
Multi-hypothesis tracking makes up the framework of the object tracker,
storing hypothesis tracks in a tree-based structure. Each tree in the framework
represents a single tracked entity. Each branch in the tree represents a
hypothetical trajectory of the given object entity - each node a tracked
detection at a specific timestamp.

\section{METHODS}
\subsection{Short-Term Tracking}
In this work, we use vision to enhance kinematic projections for improved state
estimation and uncertainty estimation while utilizing the objects' surroundings
and own appearance to predict future motion. This approach combines both visual
and kinematic aspects of aerial ISR. The dynamics of the aircraft motion, sensor
pointing, optical zoom, and target motion are modeled in an encoder-decoder
paradigm.

We utilize an attention mechanism
to allow the decoder to selectively attend to regions in the scene most important
to each projection timestep. Objects within the scene may influence the target’s
behavior at different times e.g., stationary, or non-stationary obstacles.

\begin{figure}[h]
   \begin{center}
         \includegraphics[scale=0.5]{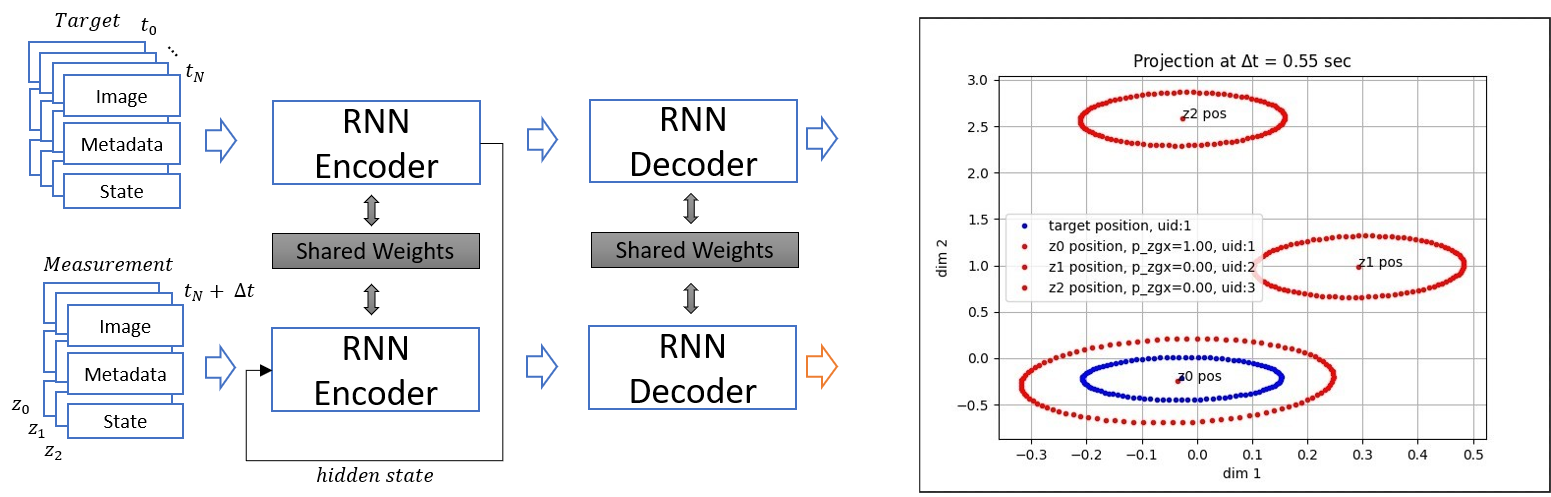}
         \caption{DeepEKF affinity example of score in the 2D latent space.}
         \label{DeepEKLatent2d}
   \end{center}
\end{figure}

DeepEKF determines the similarity between a tracklet and detection (i.e., measurement)
by computing the likelihood from the DeepEKF filter state. First the
target tracklet is encoded via the RNN Encoder, then decoded with the RNN
Decoder with Attention. The output is a $n$-dimensional projection of the target object in
latent space with a position and uncertainty ellipse at $t_N+\Delta t$. All
measurements are then passed through the same RNN Encoder and Decoder
(same weights), with the only difference being the RNN Encoder’s hidden state is
initialized from the output of the target encoding pass. The result is that each
measurement will have a state and covariance that can be compared to the target
projection. The residual position and covariance are used to compute the
likelihood value. For the 2D case, the more overlap between ellipses, the greater the likelihood
that the tracklet produced the measurement (Figure \ref{DeepEKLatent2d}).

\subsection{Long-Term Tracking (Re-identification) with Siamese Networks}
Kinematic data is not always reliable for object tracking particularly in
occluded scenes for objects that do not have well-defined kinematic behavior.
In these situations visual identification is important to ensure that multiple
tracks are not formed on a single object. Siamese networks are important for visual
identification because they are able to learn a unique latent space embedding of
an object using visual features that can be used for re-identification. Context
plays an important role in re-identification so it is important to note the
difference between short-term re-identification and long-term re-identification.
Short-term re-identification in the context of object tracking generally
realizes lower object appearance variance when compared to long-term
re-identification. It is due to this increased variance for long-term
re-identification that makes it challenging to perform in the real-world;
particularly when paired with potentially low resolution data from an aerial
perspective. One such issue is clothing changes between detections for person
re-identification since existing open-source models often times identify
features that would fail under these circumstances.

Our Siamese network architecture \cite{Chopra2005, Taigman2014} is an attention-driven model that is able to
focus on specific object features of interest to create its latent space
embedding. Our modular architecture
consists of a four basic components known as the backbone, neck, decoder, and
head. The backbone extracts visual feature maps from the image using a feature
extractor network such as a convolutional neural network (CNN) or vision
transformers (ViT). The neck acts as a translation mechanism that embeds the
feature maps from the backbone into a format suitable for the decoder. The
decoder globally attends to the feature maps to suppress irrelevant spatial
locations in the image. Finally, the head is an attention model that takes the
visual feature embedding and identifies spatial areas of interest that are used
to form the multiple final latent space embeddings.

\begin{figure}[h]
   \begin{center}
         \includegraphics[scale=0.65]{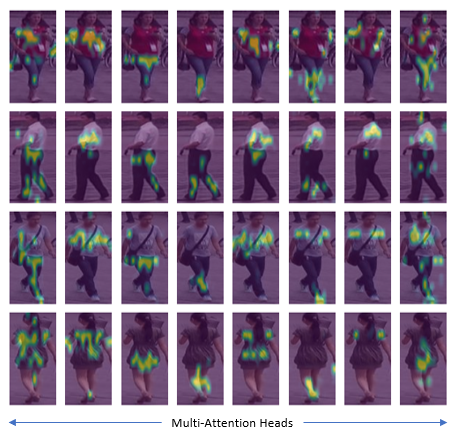}
      \caption{Attention heads for example Market-1501 images}
      \label{AttentionHeads}
   \end{center}
\end{figure}

Our models were evaluated on the open source Market-1501 dataset \cite{zheng2015market} with the official testing protocol that uses a set of query images to find the correct match across a set of gallery of images. Our attention-driven Siamese network model was able to achieve 99.4\% Rank-1 accuracy and 74.9\% mAP using this evaluation protocol. Example attention
maps for each attention head can be found in Figure~\ref{AttentionHeads}.

\begin{figure}[h]
   \begin{center}
         \includegraphics[scale=0.65]{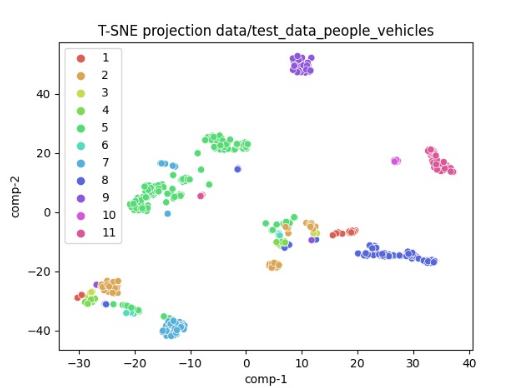}
      \caption{T-SNE plot for test Video 1. The output embeddings of the Siamese network model are clustered in latent space by an unique identifier (UID).}
      \label{tsne}
   \end{center}
   \end{figure}

We also trained and tested a similar Siamese network architecture, without an attention mechanism. This Siamese network architecture passes the image through a CNN backbone directly into an MLP head. This model was also evaluated on the Market-1501 dataset, and was able to achieve 90.7 \% Rank-1 accuracy and and
70.1\% mAP using single image comparisons. In Figure \ref{tsne}, we can visualize a t-SNE plot of the features extracted using the trained Siamese network on a sample dataset including images of ten distinct detections (7 people and 3 vehicles).

\section{EXPERIMENTS}
\subsection{Data}
\begin{figure}[h]
   \begin{center}
         \includegraphics[scale=0.35]{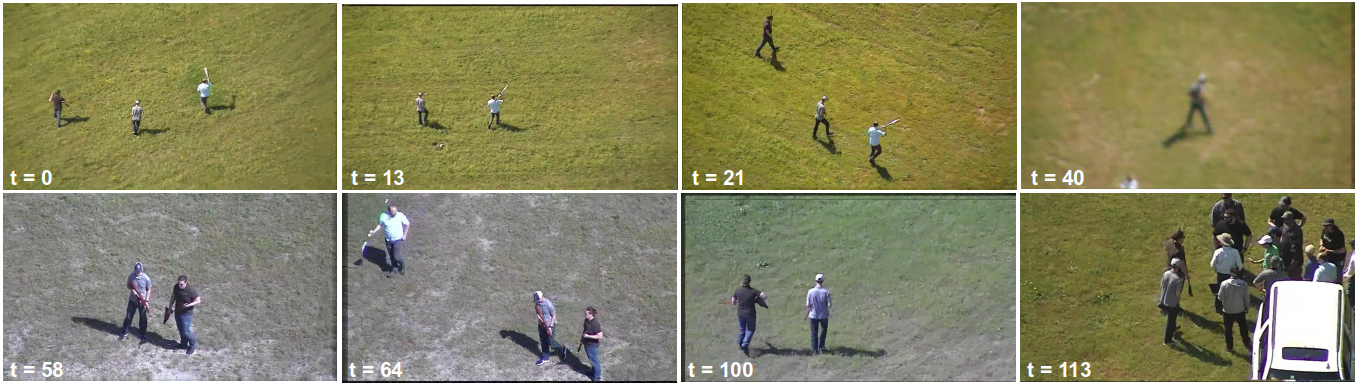}
      \caption{Sample frames from Video 1. Three people walking on an open field. Frame 0: initial frame showing the 3 people. Frame 13: person on the left with black t-shirt walked out of the scene. Frame 21: person with black t-shirt is back. Frame 40: blurred image. Frame 58: notice the changes in illumination and color. Frame 64: third person with white t-shirt is back. Frame 100: changes in point of view. Frame 113: two people join a group.}
      \label{fig:3walkers}
   \end{center}
\end{figure}

\begin{figure}[!htbp]
   \begin{center}
         \includegraphics[scale=0.35]{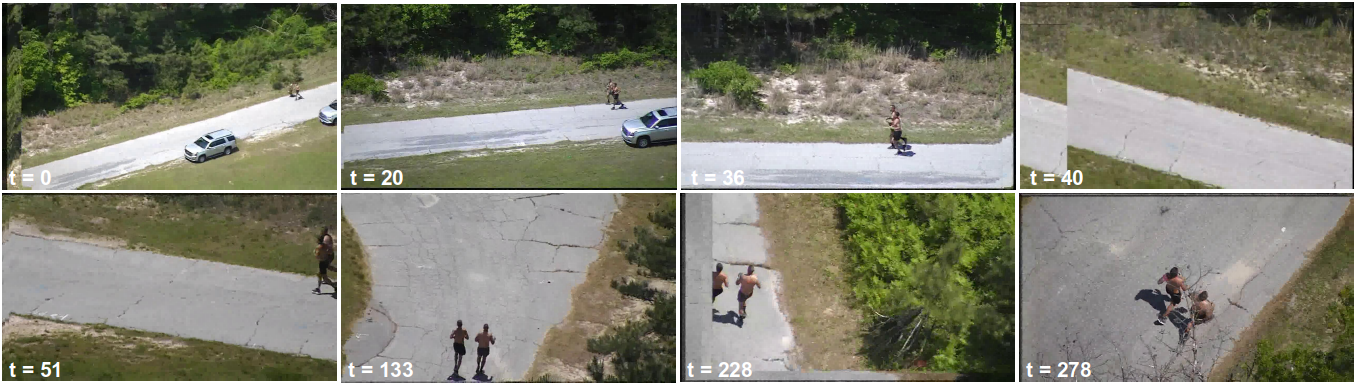}
      \caption{Sample frames from Video 2. Two people running across a road. Frame 0: initial frame showing the 2 runners. Frame 20: notice the small size of the entities. Frame 36: second runner is partially occluded and overlapping with the other runner. Frame 40: runners are out of the scene. Frame 51: runners are back after a gap of 10 frames. Frame 133: from far, the two runners are identical. Frame 228: camera movements are abrupt and cause distortions in the image. Frame 278: significant change in point of view.}
      \label{fig:2runners}
   \end{center}
\end{figure}

The data was collected from numerous flights of a fixed-wing UAS with gimbaled EO camera. The collected video was labeled using a combination of in-house and external labeling tools. We run experiments on the following two video sequences of different scenarios
involving object entities of people and vehicle, described in Table \ref{tab:videos}. In Figures \ref{fig:3walkers} and \ref{fig:2runners}, we illustrate some frames from the dataset. All images have the original size. In Video 1 (Figure \ref{fig:3walkers}), we have three people walking through a field with significant changes in scale, point of view, pose and color, with two of them disappearing and reappearing in the scene. In Video 2 (Figure \ref{fig:2runners}), we have two people running across a road. It presents significant changes in scale, and point of view, with both runners disappearing and reappearing in the scene twice along with the camera zooming in. These datasets are challenging from both kinematic and visual modeling perspectives. It has abrupt changes in positioning and relatively long periods of absence from the scene (10 frames), as well as objects with very similar appearances.

\begin{table}[ht]
\caption{Dataset description.}
\label{tab:videos}
\begin{center}
\begin{tabular}{|l|l|l|l|}
\hline
\rule[-1ex]{0pt}{3.5ex}  Video & Description & Frame Length & Kinematic Metadata \\
\hline
\rule[-1ex]{0pt}{3.5ex}  1 & three people walking in field & 124 & True \\
\hline
\rule[-1ex]{0pt}{3.5ex}  2 & two people running alongside a road & 313 & True \\
\hline
\end{tabular}
\end{center}
\end{table}

We compile together multiple trajectories for each object entity in a video
sequence to train our model’s kinematic prediction output. The dataloader reads
in image and metadata files and automatically generates sequences of varying
input and output sequence length. Each frame in the sequence contains the full
object’s characteristics including kinematic metadata (e.g. longitude, latitude,
 camera position).

\subsection{Performance Measures}
Unlike widely accepted definitions for precision and recall in the case of
classification problems, defining performance measures for object tracking is
more ambiguous. We want to precisely diagnose how well our model can accurately and
consistently keep track of distinct object entities over continuous periods of
time.

Our ground truth and predicted data come in a frame-by-frame format containing
detection information - bounding box, feature characteristics, and tracking id -
as well as kinematic metadata informing the sensor and geographic coordinate
information. Unlike bounding box coordinates and feature characteristics,
ground truth tracking id labels will most often not match predicted id labels
from the object tracker, so we must add an additional layer to determine the
optimal mapping of tracking ids.

We look at several performance measures including absence prediction, precision recall plots, longevity of tracking, etc. The primary metric we utilize to benchmark our object tracking experiments is expected average overlap (EAO), introduced in the Visual Object Tracking (VOT) challenge \cite{Kristan2015}, an averaged recall score that is measured between a range of sequence lengths defined by a kernel density estimate.

\subsection{Results and Discussion}
For each video, given the pre-computed object detections along with their bounding boxes, we applied different OT model architectures, and computed the EAO looking at consistency with the original associated track ID as well as newly created track IDs. The original image bounding boxes were resized to (100,100), and the visual components (SSD and SiameseNet models) were computed. Results are shown in Tables \ref{tab:video1-100x100} and \ref{tab:video2-100x100}. Follow the description of the used models:

\begin{itemize}
\item	Greedy: EKF (non-NN). Tracking performed using standard EKF, only based on the kinematic information of the detected objects. Pixel coordinates are used and there is no neural networks involved.
\item	MHT: EKF(pixel-based) + SSD. Kinematic score is computed using EKF estimates of pixel coordinates. Visual score is directly computed from the sum squared differences of the detected objects. 50\%/50\% weights are given to the kinematic and visual components.
\item	MHT: EKF(pixel-based) + SiameseNet. Kinematic score is computed using EKF estimates of pixel coordinates. The visual score is estimated by computing the Euclidean distance between the embedding generated using the Siamese networks. 50\%/50\% weights are given to the kinematic and visual components.
\item	MHT: EKF(pixel-based) + SiameseNet Attn. Kinematic score is computed using EKF estimates of pixel coordinates. The visual score is estimated by computing the Euclidean distance between the embedding generated using the Siamese networks with attention mechanism. 50\%/50\% weights are given to the kinematic and visual components.
\item	MHT: DEKF + SSD. Kinematic score is computed using trained DeepEKF model that estimates using visual and kinematic embeddings. Visual score is directly computed from the sum squared differences of the detected objects. 50\%/50\% weights are given to the kinematic and visual components.
\item	MHT: DEKF + SiameseNet. Kinematic score is computed using trained DeepEKF model that estimates using visual and kinematic embeddings. The visual score is estimated by computing the Euclidean distance between the embedding generated using the Siamese networks. 50\%/50\% weights are given to the kinematic and visual components.
\item MHT: DEKF + SiameseNet Attn. Kinematic score is computed using trained DeepEKF model that estimates using visual and kinematic embeddings. The visual score is estimated by computing the Euclidean distance between the embedding generated using the Siamese networks with attention mechanism. 50\%/50\% weights are given to the kinematic and visual components.
\end{itemize}

\begin{table}[ht!]
\begin{center}
   \begin{tabular}{|l|l|}
      \hline
      \rule[-1ex]{0pt}{3.5ex}  Model &
      \begin{tabular}{c} \textbf{EAO (oUID)} \\EAO (aUID)\end{tabular} \\
      \hline
      \rule[-1ex]{0pt}{3.5ex}  Greedy [EKF (non-NN)] &
      \begin{tabular}{c} \textbf{0.1750} \\0.6156\end{tabular} \\
      \hline
      \rule[-1ex]{0pt}{3.5ex}  MHT [EKF (pixel-based), SSD] &
      \begin{tabular}{c} \textbf{0.0208} \\0.8839\end{tabular} \\
      \hline
      \rule[-1ex]{0pt}{3.5ex}  MHT [EKF (pixel-based), SiameseNet] &
      \begin{tabular}{c} \textbf{0.1958} \\0.7497\end{tabular} \\
      \hline
      \rule[-1ex]{0pt}{3.5ex}  MHT [EKF (pixel-based), SiameseNet Attn] &
      \begin{tabular}{c} \textbf{0.1958} \\0.7497\end{tabular} \\
      \hline
      \rule[-1ex]{0pt}{3.5ex}  MHT [DEKF, SSD] &
      \begin{tabular}{c} \textbf{0.2881} \\0.7297\end{tabular} \\
      \hline
      \rule[-1ex]{0pt}{3.5ex}  MHT [DEKF, SiameseNet] &
      \begin{tabular}{c} \textbf{0.2256} \\0.4861\end{tabular} \\
      \hline
      \rule[-1ex]{0pt}{3.5ex}  MHT [DEKF, SiameseNet Attn] &
      \begin{tabular}{c} \textbf{0.4464} \\0.5922\end{tabular} \\
      \hline
   \end{tabular}
\end{center}
\caption{Experiments on Video 1. EAO original UID (oUID) only gives credit to the predicted track ID that was first associated with a ground truth track id. EAO any UID (aUID) gives credit to multiple associated predicted track IDs as long as there is no disagreement.}
\label{tab:video1-100x100}
\end{table}

In Table \ref{tab:video1-100x100}, the best EAO is 0.4464 for the DeepEKF using the Siamese with attention networks, followed by the DeepEKF with the standard SSD, EAO 0.2881. In this video, we have three people walking through a field with significant changes in scale, point of view, pose and color, with two of them disappearing and reappearing in the scene (as illustrated in Figure \ref{fig:3walkers}).

\begin{table}[ht!]
\begin{center}
\begin{tabular}{|l|l|}
\hline
\rule[-1ex]{0pt}{3.5ex}  Model &
\begin{tabular}{c} \textbf{EAO (oUID)} \\EAO (aUID)\end{tabular} \\
\hline
\rule[-1ex]{0pt}{3.5ex}  Greedy [EKF (non-NN)] &
\begin{tabular}{c} \textbf{0.0221} \\0.5897\end{tabular} \\
\hline
\rule[-1ex]{0pt}{3.5ex}  MHT [EKF (pixel-based), SSD] &
\begin{tabular}{c} \textbf{0.0277} \\0.8684\end{tabular} \\
\hline
\rule[-1ex]{0pt}{3.5ex}  MHT [EKF (pixel-based), SiameseNet] &
\begin{tabular}{c} \textbf{0.0371} \\0.6820\end{tabular} \\
\hline
\rule[-1ex]{0pt}{3.5ex}  MHT [EKF (pixel-based), SiameseNet Attn] &
\begin{tabular}{c} \textbf{0.0371} \\0.6694\end{tabular} \\
\hline
\rule[-1ex]{0pt}{3.5ex}  MHT [DEKF, SSD] &
\begin{tabular}{c} \textbf{0.1494} \\0.7586\end{tabular} \\
\hline
\rule[-1ex]{0pt}{3.5ex}  MHT [DEKF, SiameseNet] &
\begin{tabular}{c} \textbf{0.1790} \\0.2460\end{tabular} \\
\hline
\rule[-1ex]{0pt}{3.5ex}  MHT [DEKF, SiameseNet Attn] &
\begin{tabular}{c} \textbf{0.5045} \\0.5414\end{tabular} \\
\hline
\end{tabular}
\end{center}
\caption{Experiments on Video 2.}
\label{tab:video2-100x100}
\end{table}

In Table \ref{tab:video2-100x100}, the best EAO is 0.5045 for the DeepEKF using the Siamese with attention networks as well, followed by the DeepEKF with the standard Siamese network, EAO 0.1790. In this video, we have two people running across a road, and it presents significant changes in scale, and point of view, with both runners disappearing and reappearing in the scene twice (sample frames are illustrated in Figure \ref{fig:2runners}). This dataset is challenging for both kinematic and visual modelings. It has abrupt changes in positioning and relatively long periods of absence from the scene (8-9 frames). Additionally, the two runners are very similar in appearance and only distinguishable using the positional information.

In both experiments, our proposed DeepEKF method, using the Siamese networks with attention to generate the visual scores, outperforms the other variants. This is likely because its ability to model non-linear kinematics and re-identify objects with significant changes in pose and view.

\begin{figure}[h]
   \begin{center}
         \includegraphics[scale=0.35]{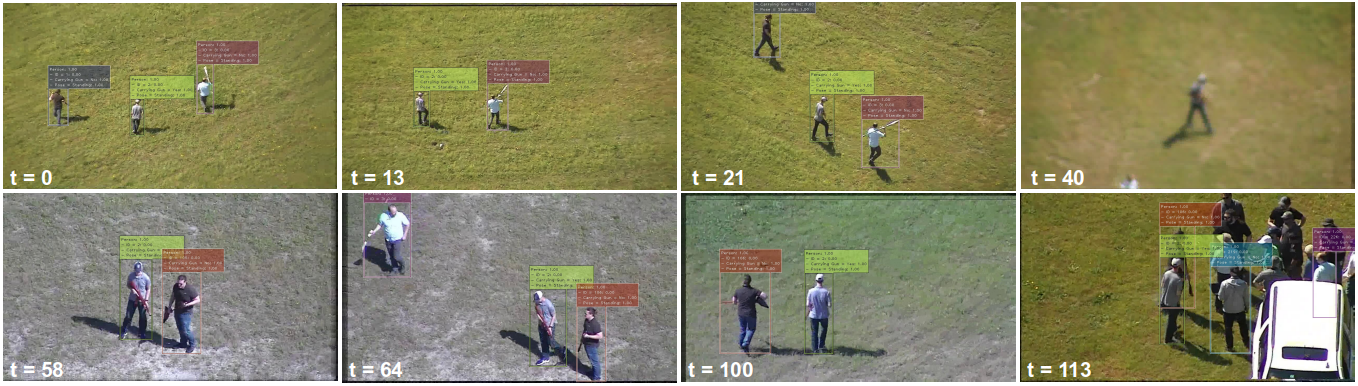}
      \caption{Tracking results from Video 1: Three people walking on an open field. Frame 0: initial tracking with 3 people. Unique identifiers (UIDs) are reported in the label box next to the identified object. Each UID is represented with different color. Frame 13: two UIDs are correctly identified. Frame 21: person with black t-shirt is back and correctly identified. Frame 40: tracking ignores poor quality detection. Frame 58: person with hat is correctly identified however a new ID is mistakenly assigned to the person with black t-shirt. Frame 64: third person with white t-shirt is back and correctly identified. Frame 100: person with black t-shirt is correctly tracked after frame 58. Frame 113: two persons continue being correctly tracked after joining a group.}
      \label{fig:3walkers_results}
   \end{center}
\end{figure}

\begin{figure}[!htbp]
   \begin{center}
         \includegraphics[scale=0.35]{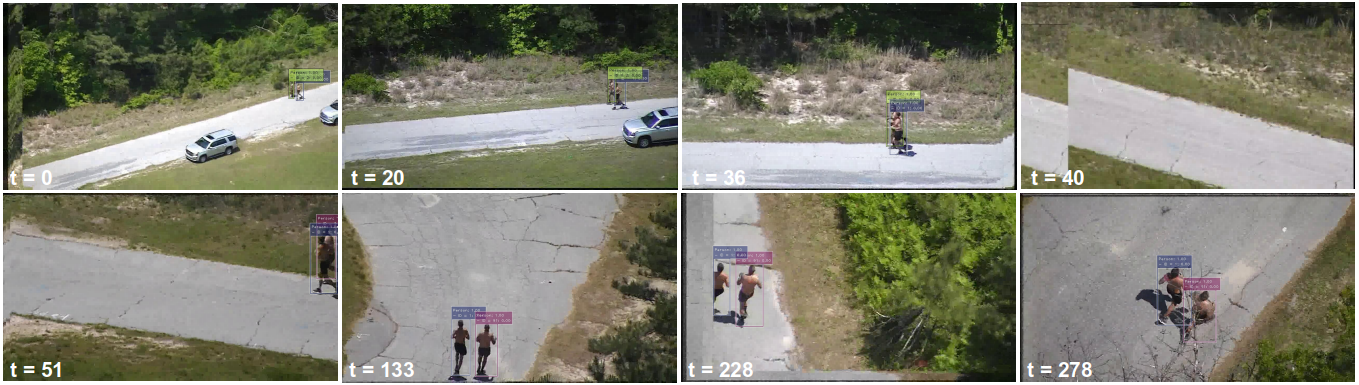}
      \caption{Tracking results from Video 2: Two people running across a road. Frame 0: initial frame showing the 2 runners. Unique identifiers (UIDs) are reported in the label box next to the identified object. Each UID is represented with different color. Frame 20: tracked objects are very close each other. Vehicle is ignored in this tracking experiment. Frame 36: runners are correctly identified even with overlapping objects. Frame 40: there is no detection but hypothesis trees are kept by the system. Frame 51: after a gap of 10 frames, the tracker correctly re-identifies the first but not the second runner. Frame 133: after the mistake at frame 51, the second runner continues being correctly tracked. Frame 228: first runner correctly tracked even with partial occlusion. Frame 278: both runners are correctly tracked until the end after frame 51.}
      \label{fig:2runners_results}
   \end{center}
\end{figure}

We use the EAO (original UID) score as our primary metric (reported in Tables \ref{tab:video1-100x100} and \ref{tab:video2-100x100}). The EAO (any UID) score does not penalize creating new track IDs for object entities, even if they have been seen before. Therefore, a model can get a high score merely by assigning a new track ID every time. However, when the EAO (original UID) score is tied, the EAO (any UID) score is useful to determine which model will make less mistakes in assigning a current UID to the wrong object entity.

\section{FUTURE WORK}
In this paper, we proposed a DeepEKF architecture for MOT in UAS applications. The method was validated on two data sets, showing improvements of the DeepEKF over the standard KF within the MHT framework. Although we report a proof of concept results, further training of the DeepEKF as well as of the Siamese networks are necessary as we collect more data. In particular, we plan to add a more extensive evaluation for the long-term tracking (re-identification) component. Another promising venue is to dynamically combine the different kinematic and visual scores within the Similarity Fuser component (Figure \ref{fig:architecture}) given the environment and track states.

\acknowledgments
We thank David Rosenbluth for providing review and feedback on this work. We thank Andrew Walsh and Kenneth Rapko Jr. for their valuable feedback on benchmarking techniques in object tracking.

% References
\bibliography{bib/report} % bibliography data in report.bib
\bibliographystyle{styles/spiebib} % makes bibtex use spiebib.bst

\end{document}